\documentclass[times,twocolumn,final,authoryear]{elsarticle}

\usepackage{ycviu}
\usepackage{framed,multirow}

\usepackage{amssymb}
\usepackage{latexsym}
\usepackage{amsmath}

\usepackage{url}
\usepackage{xcolor}
\definecolor{newcolor}{rgb}{.8,.349,.1}

\journal{Computer Vision and Image Understanding}

\begin{document}

\begin{frontmatter}

\title{End-to-End Dense Video Grounding via Parallel Regression}

\author[1]{Fengyuan \snm{Shi}} 
\author[2]{Weilin \snm{Huang}}
\author[1]{Limin \snm{Wang}\corref{cor1}}
\cortext[cor1]{Corresponding author: }
\ead{lmwang@nju.edu.cn}

\address[1]{The State Key Laboratory for Novel Software Technology, Nanjing University, Nanjing 210023, China}
\address[2]{Alibaba Group, China}

\begin{abstract}
 Video grounding aims to localize the corresponding moment in an untrimmed video given a sentence description. Existing methods often address this task in an indirect ``one-to-many'' way, i.e., predicting much more than one proposal for one sentence description, by casting it as a propose-and-match or fusion-and-detection problem. Solving these surrogate problems often requires sophisticated label assignment during training and hand-crafted removal of near-duplicate results. Meanwhile, existing works typically focus on sparse video grounding with a single sentence as input, which could result in ambiguous localization due to its unclear description. 
 In this paper, we tackle a new problem of dense video grounding, by simultaneously localizing multiple moments with a paragraph as input. From a perspective on video grounding as language-conditioned regression, we present an end-to-end parallel decoding paradigm by re-purposing a Transformer-alike architecture (PRVG). The key design in our PRVG is to use languages as queries, and regress only one temporal boundary for each sentence based on language-modulated visual representations. Thanks to its simplicity in design, our PRVG framework predicts in a ``one-to-one" manner, getting rid of complicated label assignment when training and allowing for efficient inference without any post-processing technique. In addition, we devise a robust proposal-level attention loss to guide the training of PRVG, which is invariant to moment duration and contributes to model convergence. We perform experiments on two benchmarks, namely ActivityNet Captions and TACoS, demonstrating the superiority of PRVG. We also perform in-depth studies to investigate the effectiveness of the parallel regression paradigm on video grounding.
 \end{abstract}

\end{frontmatter}

\section{Introduction}
\label{sec:intro}

\begin{figure*}
\centering
\includegraphics[width=0.85\textwidth]{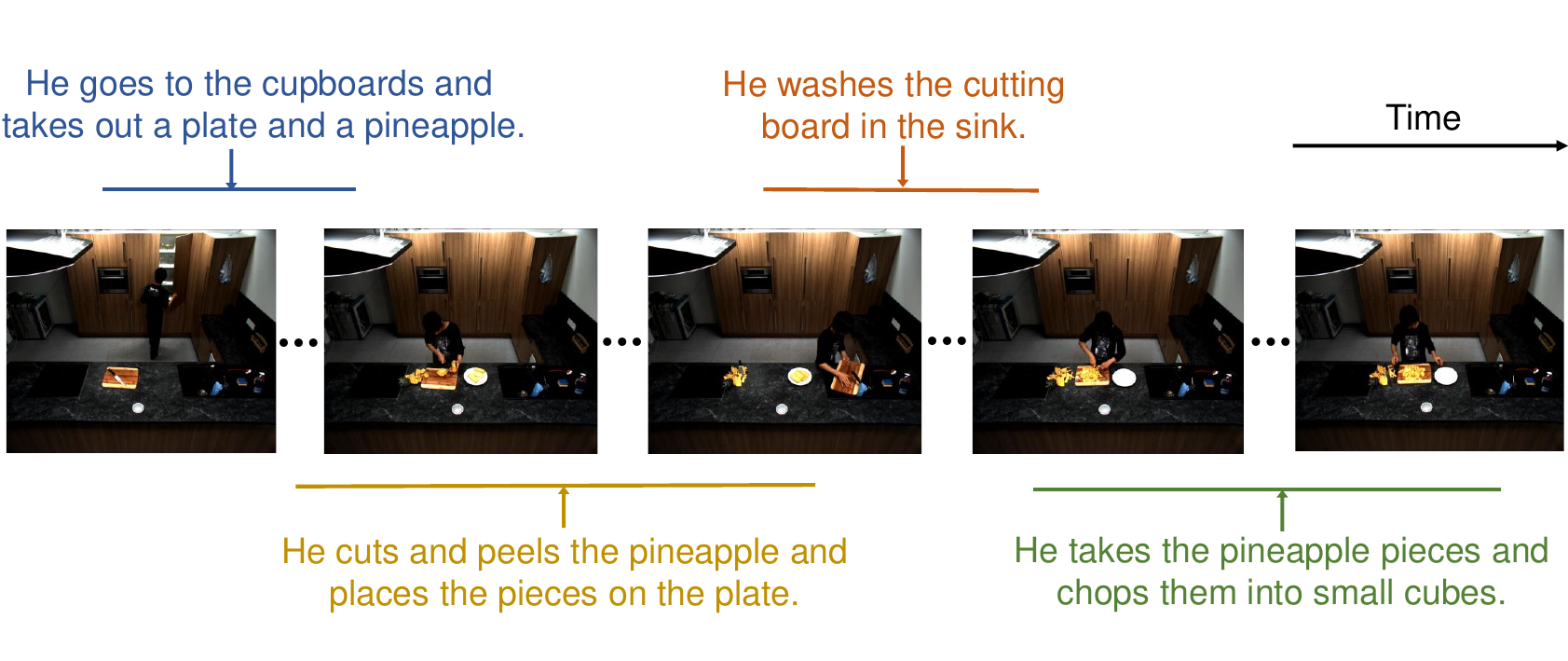}
\caption{An illustrative example of the dense video grounding task. Dense video grounding aims to jointly localize multiple temporally ordered moments described by a paragraph in an untrimmed video. }
\label{fig1}
\end{figure*}

Video Grounding (VG) is to localize the temporal moment matched with a given language description in an untrimmed video~\citep{gao2017tall,anne2017localizing,zhang2023temporal,liu2023survey}. Different from traditional computer vision tasks in video understanding such as action recognition~\citep{tran2015learning,wang2016temporal} and temporal action localization~\citep{lin2018bsn,tan2021relaxed}, which are limited to pre-defined actions, VG allows for an open set of activities with language as queries. \par

\begin{figure*}
\centering
\includegraphics[width=0.85\textwidth]{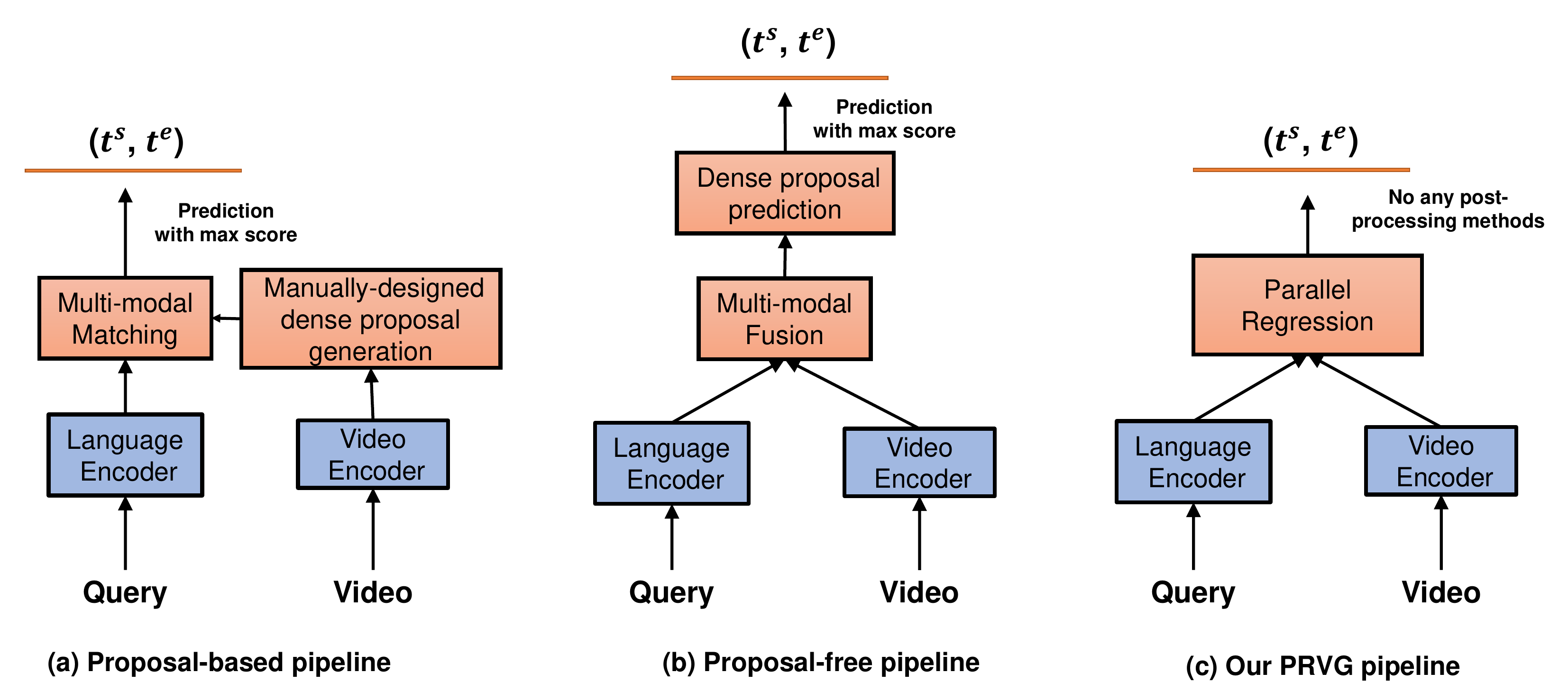}
\caption{(a) Proposal-based pipeline. (b) Proposal-free pipeline. (c) Our proposed PRVG pipeline. Both proposal-based methods and proposal-free methods address video grounding task in an indirect ``one-to-many'' manner, i.e., generating many proposals for one language description, by pre-defining dense proposals or predicting dense proposals at all locations, which suffer from complicated label assignment and near-duplicate removal. PRVG directly regresses the temporal boundary for each language description via parallel regression, without a classification branch and extra post-processing such as Ranking or NMS.}
\label{fig2}
\end{figure*}

Previous methods mainly addressed this problem in an indirect ``one-to-many'' manner, i.e., generating many proposals for one language query, by pre-defining dense proposals or predicting dense proposals at all locations, which suffer from complicated label assignment and near-duplicate removal. Specifically, they can be grouped into two categories, namely proposal-based approaches~\citep{anne2017localizing,gao2017tall,liu2018attentive,liu2018cross,xu2019multilevel,yuan2019semantic,zhang2019man,zhang2019cross,wang2020temporally,zhang2020learning,liu2021context,shin2022learning,Seol_2023_CVPR,panta2024cross} and proposal-free approaches~\citep{yuan2019find,rodriguez2020proposal,zeng2020dense,chen2020rethinking,zhang2020span,zhao2021cascaded,li2021proposal}. As shown in Fig. \ref{fig2}(a), proposal-based methods present a propose-and-match paradigm for VG task, which first manually generate a set of moment candidates using sliding window strategy or 2D temporal map~\citep{zhang2020learning}, and then calculate the similarity between each candidate with the language query. These proposals are usually of fixed length and lack flexibility to deal with various durations. It often requires large numbers of proposals to cover the ground-truth segments. In contrast, proposal-free approaches present a fusion-and-detection paradigm for VG task. As shown in Fig. \ref{fig2}(b), these methods usually first fuse two modalities of video and language and then use a dense detection framework to localize visual segments. They typically predict the offset between the current frame and boundary or the probabilities of each temporal position as the boundary, then generate dense proposals with the same or squared number of video frames. Although they can generate proposals with arbitrary length, these detection methods typically require complicated label assignment during training and removal of near-duplicate predictions. \par

Moreover, semantic information in a single sentence is limited, which might result in ambiguous temporal moments solely based on this description. Considering the example in Fig. \ref{fig1}, if we want to localize the temporal moment of ``He washes the cutting board in the sink'', this description might be unclear and lead to wrong localization, because the person may do ``washing'' several times after cutting different ingredients. However, if we know the person cuts and peels the pineapple before it, and chops the pineapple pieces into small cubes after it, the localization will be more accurate. Therefore, more contexts from other sentences are required to overcome this semantic ambiguity. However, most existing methods focus on VG task with a single sentence, which fail to capture semantic relevance and temporal clues among the multiple sentences to be grounded in a video. Recently, dense VG~\citep{bao2021dense} is introduced as a new task, and its goal is to jointly localize multiple temporal moments described by a paragraph in an untrimmed video. It proposes a DepNet to address this problem by selectively propagating the aggregated information to each single sentence. Yet, this process of aggregating temporal information of multiple sentences into a compact set will lose fine-grained context information for a specific language query. \par

To solve these problems, we present a simple yet effective framework for dense video grounding. Taking a different perspective on VG as a direct regression problem, we present an end-to-end parallel decoding paradigm by re-purposing a Transformer-alike architecture (PRVG). The key design in our PRVG framework is languages as queries, and using the paragraph to directly attend the contextualized visual representations for moments regression. This simple design in our parallel decoding paradigm shares two critical advantages. First, the cross-attention between languages and visual representations endows PRVG with the global context view, which provides the flexibility of accurate localization for temporal moments of various durations at different locations. Second, the self-attention in sentences enables PRVG to model the mutual context among different sentences, which could be leveraged to overcome the semantic ambiguity of each individual sentence. In addition, we also design a robust proposal-level attention loss to explicitly guide the attention mechanism training in PRVG, resulting in a more accurate regression model. Due to its simplicity in design, our PRVG can be deployed efficiently without any complicated label assignment and post-processing technique. 

We perform extensive experiments on two benchmarks of ActivityNet Captions~\citep{krishna2017dense} and TACoS~\citep{regneri2013grounding}, demonstrating the superiority of PRVG. We also perform in-depth ablation studies to investigate the effectiveness of parallel regression scheme in our PRVG and try to present more insights on the task of dense VG. Our main contributions are summarized as follows: 
\begin{itemize}
    \item We cast VG as a direct regression problem and present a simple yet effective framework (PRVG) for dense VG. PRVG directly regresses the temporal boundary for each language query to the final result, leading to accurate and efficient inference.
    \item We design a robust and scale-invariant proposal-level attention loss function to guide the training of PRVG for better performance. 
    \item \textcolor{black}{We analyze the differences between object detection and (dense) video grounding, and adapt DETR for (dense) VG from the view of query design. We make a detailed discussion and comparison with DETR and DETR-based (dense) video grounding methods. We hope our paper can provide readers with some insights about how to extend the success of DETR, such as query designs and multimodal interaction.}
    \item Extensive experiments demonstrate the superiority of PRVG and the effectiveness of parallel decoding paradigm on dense video grounding task.
\end{itemize}

\section{Related Work}
\noindent \textbf{Video Grounding.}
Video grounding~\citep{gao2017tall,anne2017localizing,zhang2023temporal,liu2023survey} aims to localize a temporal moment given by a sentence description. Existing methods can be grouped into two categories, including proposal-based approaches and proposal-free approaches.
{\em Proposal-based approaches} cast this task as a cross-modal matching problem and solve it in a ``propose-and-match'' manner. Early works~\citep{gao2017tall,anne2017localizing,liu2018attentive,liu2018cross} adopted sliding window strategy to generate proposals and produced matching scores between the language query and each proposal. The proposal with the highest matching score was selected as the final prediction result. To generate more precise temporal moments,~\citep{chen2018temporally,xu2019multilevel,yuan2019semantic,zhang2019cross,liu2020jointly} regressed temporal boundaries on pre-defined multi-scale anchors. MAN~\citep{zhang2019man} produced proposals of different lengths by temporal pooling and 1D convolution. Zhang \textit{et al.}~\citep{zhang2020learning} proposed a Temporal Adjacent Network to model the temporal relations between video moments by a two-dimensional map. Subsequent works follow 2D-TAN~\cite{zhang2020learning} and localize the moments on the 2D proposal maps~\citep{liu2021context,shin2022learning,Seol_2023_CVPR,panta2024cross}. CBLN~\citep{liu2021context} proposed to score all pairs of start and end indices simultaneously by a biaffine mechanism. BMRN~\citep{Seol_2023_CVPR} proposed a novel boundary matching and refinement mechanism to obtain variable boundaries from the 2D temporal proposal map. 

To produce more flexible results with arbitrary length, various {\em proposal-free} approaches were proposed~\citep{rodriguez2020proposal,zeng2020dense,chen2020rethinking,zhang2020span,zhao2021cascaded}, which predicted the probabilities of each frame as start and end boundaries, and then got a two-dimensional score map whose element indicates the predicted probability of each moment candidate. \citep{rodriguez2020proposal} used sentence features as dynamic filters for accurate attention. To eliminate the problem of imbalance between positive and negative examples, DRN~\citep{zeng2020dense} used the distances between the frame within the ground truth and the starting (ending) frame as dense supervision. CPN~\citep{zhao2021cascaded} constructs a segment-tree-based structure to generate grounding results in a cascaded manner.

These methods addressed VG with indirect surrogate problems, i.e., predicting much more than one proposal for one sentence description which requires sophisticated label assignment during training and hand-crafted removal of near-duplicate results. On the contrary, we cast VG as a direct regression problem, and regress only one moment for each sentence description without the need of the classification head, avoiding sophisticated label assignment and post-processing.

\noindent \textbf{Dense Video Grounding.}
Bao \textit{et al.}~\citep{bao2021dense} introduced the task of dense video grounding, and its goal is to jointly localize all temporal moments corresponding to these sentence descriptions, given an untrimmed video and a paragraph of sentence descriptions. To extract temporal and semantic information of sentence descriptions, Bao \textit{et al.}~\citep{bao2021dense} proposed a Dense Events Propagation Network (DepNet) in an aggregation-and-propagation scheme. Compared to our intra-modal and inter-modal context modeling in our PRVG, DepNet missed fine-grained context information for a specific language query. SVPTR~\citep{Jiang_2022_CVPR} proposed a contrastive encoder to learn video paragraph alignment and adopted semi-supervised learning to reduce the requirements of labeled data. HSCNet~\citep{Tan_2023_CVPR} developed a hierarchical structure to enhance semantics-aligned representations at different levels. 

\textcolor{black}{SVPTR, HSCNet and our PRVG all introduce hierarchical attention designs, while there are some differences. First, SVPTR and HSCNet mainly focus on attention designs in model architectures, while our proposed PRVG makes more efforts on attention loss for model training. Second, both SVPTR and HSCNet introduce hierarchical structures according to the text inputs (words, sentences and paragraphs), while our PRVG focuses on the video side and proposes an attention loss extending from a video clip to a moment proposal.}

\begin{figure*}[t]
\centering
\includegraphics[width=1.0\textwidth]{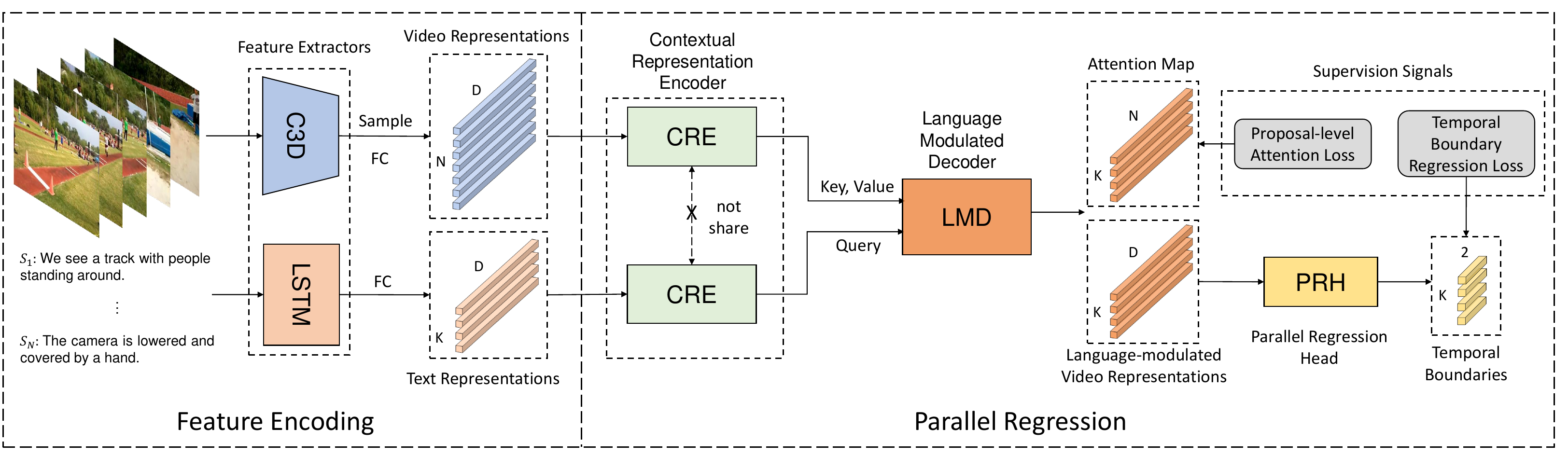}
\caption{{\bf Pipeline of PRVG}. Our PRVG streamlines the process of dense video grounding with a direct and parallel decoding paradigm, which is composed of two steps: feature encoding and parallel regression.
In feature encoding phrase, we extract the features of video and sentences by 3D CNN and LSTM, respectively.  As for parallel regression, a Contextual Representation Encoder (CRE) is proposed to augment the feature representations with global context information for both modals, and a Language Modulated Decoder (LMD) using languages as queries coupled with a Parallel Regression Head (PRH) is proposed to directly predict the temporal moment for each sentence descriptions. Our PRVG is able to  capture intra-modal global structure information for contextualized representation and model the cross-modal relation in a global view for flexible and accurate moment localization.}
\label{fig3}
\end{figure*}

\noindent \textbf{Vision Transformer and DETR.}
Transformer was first presented by~\citep{vaswani2017attention} in machine translation task. Due to its powerful global temporal modeling capability, transformer is applied in many other NLP tasks, such as named entity recognition~\citep{yan2019tener} and language understanding~\citep{devlin2018bert}. Inspired by the recent advances in NLP tasks, transformer is applied in computer vision, such as image classification~\citep{dosovitskiy2020image,liu2021swin,he2022masked,touvron2021training}, object detection~\citep{carion2020end,zhu2020deformable} and action proposal generation~\citep{tan2021relaxed}. 

\textcolor{black}{DETR is a representative transformer-based object detector, performing bipartite matching to choose queries matched with ground truth from the query set. And there are some works like Moment-DETR~\citep{lei2021detecting} and MH-DETR~\citep{xu2023mh} using DETR for video moment retrieval. Our proposed PRVG is also built on DETR. The differences are that we use the languages as queries, while they typically use learnable positional embeddings as queries which are fixed once the training is finished. Our improvements in decoder design enable our framework more flexible and explainable. Moreover, benefiting from the one-to-one correspondence between the query and target moment, we can introduce a proposal-level attention loss to further improve the performance, by allowing the model to attend to the ground truth region, determine the salient temporal locations, and suppress attention values at locations less important in the ground truth region.}

\section{Method}

We formulate the dense video grounding task as follows. Given an untrimmed video $V$ and a paragraph of $K$ temporally ordered sentences $\{S_1, \cdots, S_K\}$. The goal of dense VG is to jointly localize multiple temporal moments $\{T_1, \cdots, T_K\}$ for all these sentences, where $T_k = (t_k^s, t_k^e)$. $t_k^s$ and $t_k^e$ denote the start and end time points of the temporal moment corresponding to the $k$-th sentence, respectively. Specifically, the video is represented by a sequence of frames, i.e. $V = \{f_t\}_{t=1}^{L_V}$, where $f_t$ is the $t$-th frame and $L_V$ is the total number of frames. The $k$-th sentence $S_k = \{s_{ki}\}_{i=1}^{L_{S_k}}$, where $s_{ki}$ is the $i$-th word in the sentence and $L_{S_k}$ is the total number of words. \par

As shown in Fig. \ref{fig3}, the pipeline of our proposed PRVG consists two steps: feature encoding and parallel regression. First, we extract features of video and sentence descriptions in the feature encoding phrase. In the parallel regression phrase, we use Contextual Representation Encoder (CRE) to model global context and get contextual representations for both modals, then a Language Modulated Decoder (LMD) is adapted to aggregate the visual information from the entire video by using the language representations as queries. Finally, a Parallel Regression Head (PRH) directly predicts the temporal boundaries for all sentence descriptions in the paragraph simultaneously without extra post-processing such as Ranking or NMS.

\subsection{Preliminary}
Parallel regression module in our proposed PRVG is built on transformer~\citep{vaswani2017attention}. We brieﬂy review the vanilla transformer here. Transformer has an encoder-decoder structure, where both encoder and decoder are composed of a stack of $L$ identical layers consisting of multi-head attention and position-wise feed-forward network sub-layers. For each two sub-layers, a residual connection~\citep{he2016deep} is employed, followed by layer normalization~\citep{ba2016layer}. \par
Attention mechanism is the keypoint of transformer. Given query embedding $Q$, key embedding $K$ and value embedding $V$, the scaled dot-product attention is computed as:
\begin{equation}
    \text{Attention}(Q,K,V) = \text{softmax}(\frac{QK^\mathrm{T}}{\sqrt{d_k}})V \label{eq0}
\end{equation}
For self-attention, they are projections from the same input, while for cross-attention, they represent projections from different modalities. In Transformer, multi-head attention is the concatenation of multiple single attention head outputs.

\begin{equation}
\begin{aligned}
    \text{MHA}(Q,K,V) &= \text{Concat}(\text{head}_1, \cdots, \text{head}_H)W^O, \\
     \text{head}_i &= \text{Attention}(QW_i^Q, KW_i^K, VW_i^V),
\end{aligned}
\end{equation}
where $W$ are learnable projection matrices for feature transformation.

\subsection{Feature Encoding}
For fair comparisons with previous methods~\citep{zhang2020learning,bao2021dense}, we use the same video and sentence feature encoding. Specifically, for each word $s_{ki}$ in the input sentence $S_k$, we first generate its word embedding vector $w_{ki} \in R^d$ by the Glove word2vec ~\citep{pennington2014glove}, then feed the word embedding sequence into a three-layer LSTM network~\citep{hochreiter1997long}. We use the last hidden state $f^{S_k} \in R^{D}$ as the feature representation of the input sentence $S_k$, and get the sentence features $F^S \in R^{K \times D}$. For an input sequence of video frames, we divide them into $N_v$ small clips without overlap, where each clip contains $L$ frames. For each clip $v_i$, we use a pre-trained C3D Network~\citep{tran2015learning} to extract its feature. Then we feed them into a feed-forward neural network to obtain projected features $f_{v_i} \in R^{D}$, which has the same dimension as sentence feature. For fast processing, we uniformly sample $N$ clips~\citep{wang2016temporal} from each video and get video features $F^V \in R^{N \times D}$.

\subsection{Parallel Regression}
\textcolor{black}{We introduce transformer into dense video grounding task, and extend it to the temporal dimension for video and paragraph modeling.} Specifically, our parallel regression contains a Contextualized Representation Encoder (CRE), a Language Modulated Decoder (LMD), and a customized Parallel Regression Head (PRH). CRE adopts the self-attention operations for capturing long-range dependency between visual features to enhance its visual representation power and extracting semantic relevance among language queries grounded in a paragraph. The LMD directly uses languages as queries to globally attend and aggregate encoder output for obtaining compact language-modulated visual representation. This encoder-decoder architecture constitutes a coherent pipeline to perform language modulated aggregation over visual features for subsequent parallel regression. Finally, the parallel regression head accomplishes parallel regression based on this language-modulated visual representations.

\subsubsection{Contextualized Representation Encoder}
As shown in Fig. \ref{fig4} (a), CRE consists of a multi-head self-attention layer and feed forward network with a residual connection~\citep{he2016deep}. As attention mechanism is permutation-invariant, we add the positional embedding to the input features. 
The calculation can be formulated as:
\begin{footnotesize}
\begin{equation}
\begin{aligned}
    & Q^{CRE} = K^{CRE} = V^{CRE}  = F, \\
    & F^{CRE^{'}} = \text{LN}(Q^{CRE} + \text{MHA}(Q^{CRE}, K^{CRE}, V^{CRE})), \\
    & F^{CRE} = \text{LN}(F^{CRE^{'}} + \text{FFN}(F^{CRE^{'}})),
\end{aligned}
\end{equation}
\end{footnotesize}where LN$(\cdot)$ denotes layer normalization~\citep{ba2016layer}, MHA$(\cdot)$  means multihead attention and FFN$(\cdot)$ is the feed forward network. For sentences, the input is $F^S \in R^{K \times D}$, and the output is $F^{CRE}_{L}$. For video, the input is $F^V \in R^{N \times D}$, and the output is $F^{CRE}_V$.
This contextualized representation encoder enables our video features to have a global view by using self-attention operations over the entire video sequence and the resulted contextualized visual representations. Besides, CRE extracts the temporal clues and semantic relevance among the sentences in the paragraph and facilitates the contextualized language representations.

It's worth noting that there is a large semantic gap between video and language representation, thus two CREs that do not share parameters are adopted to keep as much useful information as possible for each of the two modals. 

\subsubsection{Language Modulated Decoder}
\begin{figure}[t]
\centering
\includegraphics[width=0.85\columnwidth]{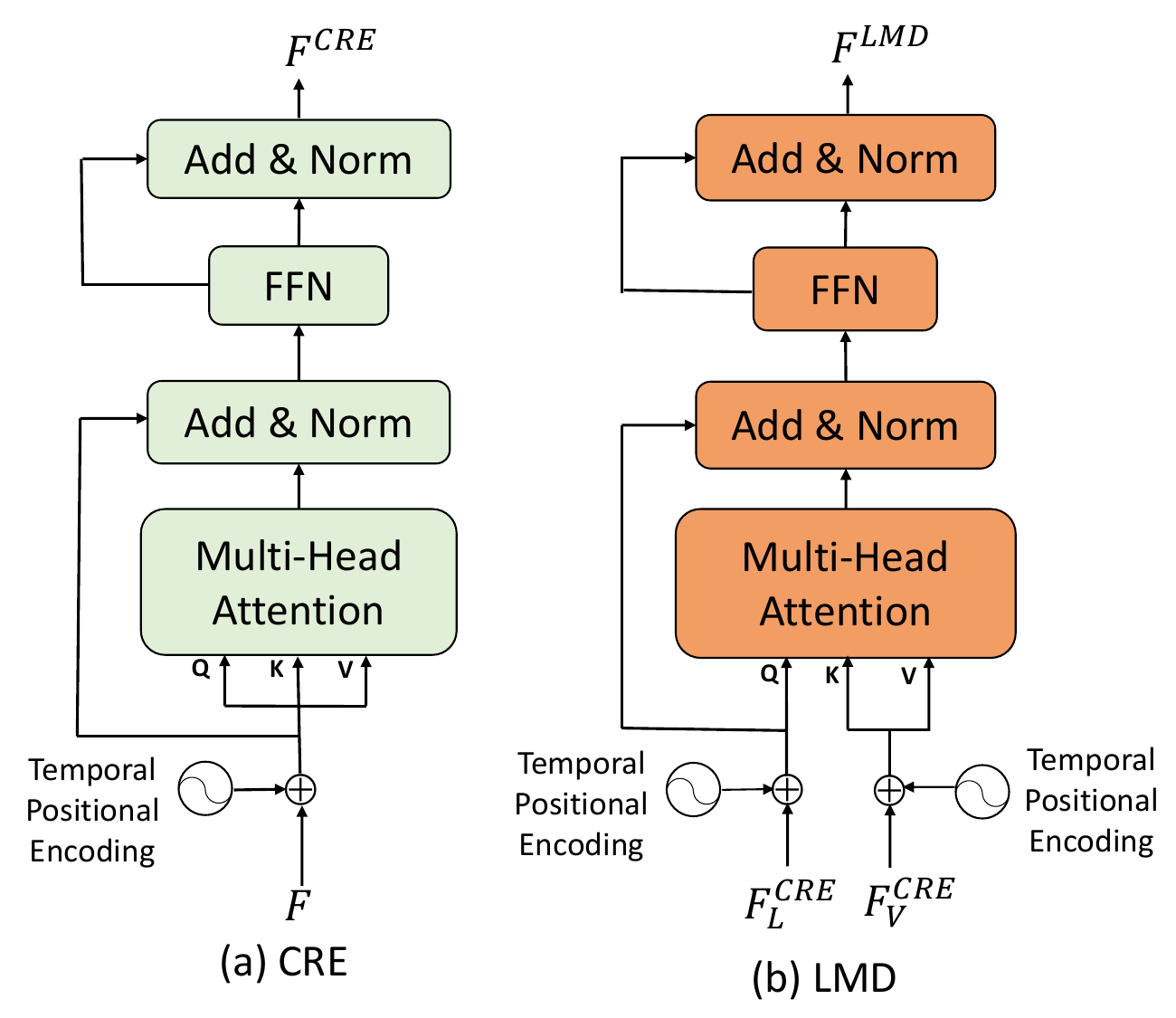} 
\caption{(a) Contextualized Representation Encoder (CRE). (b) Language Modulated Decoder (LMD). CRE is built on self-attention mechanism, whose input is the video or sentence features. CRE models long-term dependencies in video and extracts semantic relevance among the sentences in a paragraph, facilitating to contextualized representations for both modals. LMD is built on cross-attention mechanism, using language as queries and video features as keys and values. LMD aggregates visual information under the guidance of language queries for subsequent parallel regression.}
\label{fig4}
\end{figure}

As shown in Fig. \ref{fig4} (b), LMD adopts a similar architecture to CRE, which is composed of a multi-head cross-attention layer and feed forward network with a residual connection. More specifically, we also add positional encoding to the sentence features. The input of the decoder is $K$ sentence features $f^S = \{f^{S_i}\}_{i=1}^K$ and the encoder's output $E_{enc}$. The calculation can be formulated as:
\begin{footnotesize}
\begin{equation}
\begin{aligned}
    & Q^{LMD} = F_L^{CRE}, \ K^{LMD} = V^{LMD}  = F_V^{CRE}, \\
    & F^{LMD^{'}} = \text{LN}(Q^{LMD} + \text{MHA}(Q^{LMD}, K^{LMD}, V^{LMD})), \\
    & F^{LMD} = \text{LN}(F^{LMD^{'}} + \text{FFN}(F^{LMD^{'}}))
\end{aligned}
\end{equation}
\end{footnotesize}The output of the decoder is language-modulated visual representations $X^{LMD} \in R^{K \times D}$.

\textcolor{black}{Different from using object queries represented by learnable positional embeddings to decode detected objects in the decoder of DETR, our proposed PRVG uses languages as queries. Using languages as queries endows several advantages, compared to the straightforward application of DETR to video grounding, including flexibility with variable queries, no need for negative samples, and interpretability of language queries. This more explainable and flexible modeling capacity enables PRVG to deal with the task of dense VG in an open-set setting. Detailed discussions between PRVG and DETR in the later subsection.}

\subsubsection{Parallel Regression Head} 
Given the LMD's output $F^{LMD}$, the parallel regression head consisting of a 3-layer FFN with ReLU activation function directly predicts the temporal boundary ($t^s$, $t^e$) for each sentence in parallel. It is worth noting that PRVG generates only one prediction result for each language query, which is more efficient than one-to-many prediction methods.

\subsection{Discussion with DETR}

\textcolor{black}{
DETR~\citep{carion2020end} views object detection as a direct set prediction problem and uses $N$ object queries represented by \textbf{learnable positional embeddings} to decode detected objects. It is intuitive and easy to transfer DETR to (dense) video grounding, by just taking the query with the highest confidence as the target moment, as shown in Figure 5.
Based on the fused representation of video and sentence, we can propose DETR architecture to solve this grounding problem. However, the task characteristics and complexity of these two tasks are not the same, and it is non-trivial to adapt DETR to (dense) VG. We argue that this straightforward application of DETR still tackles (dense) VG in an indirect “one-to-many”,  and has some flaws compared with using \textbf{languages as queries} in our PRVG. We will make a detailed discussion with DETR and DETR-based VG methods next, and the experimental comparisons of PRVG and DETR-based models are also provided in Section \ref{section:detr}.
}

\noindent\textbf{Flexibility with variable queries.}
First, the number of the queries in the decoder of PRVG is variable, but that of DETR is fixed. For dense VG task, there can be different numbers of sentences in different input paragraphs, which makes it hard for DETR-based models with a fixed number of queries to expand to the dense VG task. Although we can adapt DETR-based models to dense VG with post-processing methods such as beam search, the temporal relationship and semantic relevance among the temporally ordered sentences in the paragraph cannot be fully explored. However, our proposed PRVG can deal with any number of sentences by the same model and fully model the temporal interactions among the sentences by self-attention in the decoder. Second, our language queries are adjusted adaptively with different input sentences and can be generalized to unseen sentences during testing, while theirs are fixed once the training is finished, and the same for different inputs. This fixed query scheme lacks the flexibility to deal with an open set of activities and might lead to performance drop in case of semantics shift.

\noindent\textbf{Retrieval framework without negative samples.}
DETR was introduced in object detection, which is essentially different from our video grounding in terms of task complexity. VG and dense VG both aim to localize the temporal moment for a sentence description from the whole video, and can be regarded as a {\bf retrieval} task. Compared with traditional detection, the search space is bigger and the imbalance of positive and negative examples is more serious if we transform retrieval task into detection. Therefore, it is hard to train a DETR-based VG model which depends on a classification branch to distinguish between foreground and background. Thanks to our design of languages as queries and parallel regression regime, PRVG directly predicts the final result for each sentence description in a one-to-one matching between queries and ground truth without classification, and there is no need for negative samples in the training process, thus eliminating the extreme foreground-background class imbalance.

\noindent\textbf{Interpretability of language queries.}
Each query in our decoder represents the concrete semantics of a sentence while the learned moment queries in DETR-based models are optimized in a data-driven manner without clear meaning. The clear semantics of queries makes our model easy to understand and generalize. Besides, each query in their decoder can predict both foreground and background, depending on the input. As the semantics of the queries of DETR-based models are not clear and the matching between the queries and ground truth moment is uncertain, we cannot explicitly supervise the attention weight training with ground truth region. In contrast, our proposed PRVG uses languages as queries and each query directly regresses the corresponding temporal boundary. So we can use an attention loss to guide the training of PRVG for better performance.

\subsection{Training}
To optimize the proposed model, we adopt a multi-task loss $\mathcal{L}$ including temporal boundary regression loss $\mathcal{L}_{reg}$ and a robust proposal-level attention loss $\mathcal{L}_{attn}$:
\begin{equation}
    \mathcal{L} = \mathcal{L}_{reg} + \mathcal{L}_{attn}. \label{eq1}
\end{equation}

\noindent \textbf{Temporal boundary regression loss.}
A training example consists of an input video $V$, a paragraph of language queries $\{S_1, S_2, \cdots, S_K\}$, and normalized temporal annotations $\{T_1, T_2, \cdots, T_K\}$. The predicted results are denoted as $\{\hat{T}_1, \hat{T}_2, \cdots, \hat{T}_K\}$. We follow~\citep{carion2020end} and adapt its 2D bounding box loss to temporal 1D form. Then we get temporal boundary regression loss as follows:
\begin{equation}
    \mathcal{L}_{reg} = \frac{1}{K} \sum_{i=1}^K (\lambda ||T_i - \hat{T}_i||_1 + \beta \mathcal{L}_{iou}(T_i, \hat{T}_i) ),\label{eq2}
\end{equation}
where $\lambda$ and $\beta$ are trade-off hyper-parameters.

\noindent \textbf{Proposal-level attention loss.}
To improve the accuracy of the learned sentence-to-video attention, ABLR~\citep{yuan2019find} designs a position-wise attention loss to encourage the video clips within the ground truth window to have higher attention values:
\begin{equation}
    \mathcal{L}_{attn} = -\frac{1}{K} \sum_{j = 1}^K \frac{\sum_{i=1}^N \hat{a}_i^j \log(a_i^j)}{\sum_{i=1}^N \hat{a}_i^j}, \label{eq3}
\end{equation}
where $a_i^j$ is the attention weight between the $i$-th clip and $j$-th sentence, and $\hat{a}_i^j = 1$ if the $i$-th clip is in the ground truth window, else $\hat{a}_i^j = 0$. 

We argue that this attention loss is not robust since it considers all temporal positions in the ground truth independently rather than as a whole. First, different positions in the ground truth might not equally contribute to the final regression and we should not equally constrain the attention at all locations to be very high.  Besides, Equation (\ref{eq3}) will not be zero and gradients will continue to backpropagate, even if the model can attend to the described moment accurately due to $\sum_{i=1}^N a_i^j = 1$, which is easy to lead to over-fitting of the model and results in inferior results. 

Therefore, we devise a robust proposal-level attention loss, by relaxing the position-wise constraint. We only require the sum of attention weights in ground truth should be high. The loss function is defined as:
\begin{equation}
    \mathcal{L}_{attn} = -\frac{1}{K} \sum_{j = 1}^K \log( \sum_{i = t_j^s}^{t_j^e} a_i^j). \label{eq4}
\end{equation}
Compared with position-wise attention loss, our proposed proposal-level attention loss regards the proposal as a whole. It not only makes the model attend to the ground truth region, but also allows the model itself to further determine the high-salient temporal locations and suppress attention values at locations less important in the ground truth region for VG. Besides, the new attention loss is scale-invariant to the moment length, which is robust for regression and helps to converge.

\begin{table*}[t]
\centering
\scalebox{0.9}{
\begin{tabular}{lcccccccc}
\hline
\multicolumn{1}{l|}{\multirow{2}{*}{Method}}                          & \multicolumn{4}{c|}{ActivityNet Captions}                                              & \multicolumn{4}{c}{TACoS}                                         \\
\multicolumn{1}{l|}{}                                                 & 0.3            & 0.5            & 0.7            & \multicolumn{1}{c|}{mIoU}           & 0.1            & 0.3            & 0.5            & mIoU           \\ \hline
\multicolumn{9}{l}{\textit{Video Grounding Methods}}                                                                                                                                                                              \\ \hline
\multicolumn{1}{l|}{MCN~\citep{anne2017localizing}}   & 39.35          & 21.36          & 6.43           & \multicolumn{1}{c|}{15.83}          & 14.42          & -              & 5.58           & -              \\
\multicolumn{1}{l|}{CTRL~\citep{gao2017tall}}         & 47.43          & 29.01          & 10.34          & \multicolumn{1}{c|}{20.54}          & 24.32          & 18.32          & 13.30          & 11.98          \\
\multicolumn{1}{l|}{TGN~\citep{chen2018temporally}}   & 43.81          & 27.93          & 11.86          & \multicolumn{1}{c|}{29.17}          & 41.87          & 21.77          & 18.90          & 17.93          \\
\multicolumn{1}{l|}{QSPN~\citep{xu2019multilevel}}    & 52.13          & 33.26          & 13.43          & \multicolumn{1}{c|}{-}              & 25.31          & 20.15          & 15.20          & -              \\
\multicolumn{1}{l|}{ABLR~\citep{yuan2019find}}        & 55.67          & 36.79          & -              & \multicolumn{1}{c|}{36.99}          & 34.70          & 19.50          & 9.40           & 13.40          \\
\multicolumn{1}{l|}{DRN~\citep{zeng2020dense}}        & -              & 45.45          & 24.36          & \multicolumn{1}{c|}{-}              & -              & -              & 23.17          & -              \\
\multicolumn{1}{l|}{2D-TAN~\citep{zhang2020learning}} & 59.45          & 44.51          & 26.54          & \multicolumn{1}{c|}{-}              & 47.59          & 37.29          & 25.32          & -              \\
\multicolumn{1}{l|}{VSLNet~\citep{zhang2020span}}     & 63.16          & 43.22          & 26.16          & \multicolumn{1}{c|}{43.19}          & -              & 29.61          & 24.27          & 24.11          \\
\multicolumn{1}{l|}{CPNet~\citep{li2021proposal}}     & -              & 40.56          & 21.63          & \multicolumn{1}{c|}{40.65}          & -              & 42.61          & 28.29          & 28.69          \\
\multicolumn{1}{l|}{BPNet~\citep{xiao2021boundary}}   & 58.98          & 42.07          & 24.69          & \multicolumn{1}{c|}{42.11}          & -              & 25.96          & 20.96          & 19.53          \\
\multicolumn{1}{l|}{CBLN~\citep{liu2021context}}      & 66.34          & 48.12          & 27.60          & \multicolumn{1}{c|}{-}              & 49.16          & 38.98          & 27.65          & -              \\ 
\multicolumn{1}{l|}{TACI~\citep{shin2022learning}}      & -          & 45.50          & 27.23          & \multicolumn{1}{c|}{-}              & -          & -          & -          & -              \\ 
\multicolumn{1}{l|}{BMRN~\citep{Seol_2023_CVPR}}      & 66.34          & 48.47          & 31.15          & \multicolumn{1}{c|}{-}              & -          & -          & -          & -              \\ \hline
\multicolumn{9}{l}{\textit{Dense Video Grounding Methods}}                                                                                                                                                                        \\ \hline
\multicolumn{1}{l|}{BS~\citep{bao2021dense}}          & 62.53          & 46.43          & 27.12          & \multicolumn{1}{c|}{-}              & 48.46          & 38.14          & 25.72          & -              \\
\multicolumn{1}{l|}{3D-TPN~\citep{zhang2020learning}} & 67.56          & 51.49          & 30.92          & \multicolumn{1}{c|}{-}              & 55.05          & 40.31          & 26.54          & -              \\
\multicolumn{1}{l|}{DepNet~\citep{bao2021dense}}      & 72.81          & 55.91          & 33.46          & \multicolumn{1}{c|}{-}              & 56.10          & 41.34          & 27.16          & -              \\
\multicolumn{1}{l|}{SVPTR~\citep{Jiang_2022_CVPR}}      & 78.07          & 61.70          & 38.36          & \multicolumn{1}{c|}{55.91}              & 67.91          & 47.89          & 28.22          & 31.42              \\
\multicolumn{1}{l|}{HSCNet~\citep{Tan_2023_CVPR}}      & \textbf{81.89}          & \textbf{66.57 }         & \textbf{44.03}          & \multicolumn{1}{c|}{\textbf{69.71}}              & \textbf{76.28}          & \textbf{59.74}          & \textbf{42.00}          & \textbf{40.61}              \\
\multicolumn{1}{l|}{PRVG (Ours)}                                & 78.27 & 61.15 & 37.83 & \multicolumn{1}{c|}{55.62} & 61.64 & 45.40 & 26.37 & 29.18 \\ \hline
\end{tabular}
}
\caption{Performance comparison on ActivityNet Captions and TACoS. }
\label{table1}
\end{table*}

\section{Experiment}
\subsection{Datasets and Evaluation Metrics}
\noindent\textbf{ActivityNet Captions}~\citep{krishna2017dense} consists of 19,209 untrimmed videos of open-domain activities, which is originally built for dense-captioning events task and recently introduced for video grounding. For a fair comparison, we follow the experimental setting in~\citep{zhang2019cross}, using val\_1 as validation set and val\_2 as testing set, which have 37,417, 17,505, and 17,031 moment-sentence pairs for training, validation, and testing, respectively. The average duration of the videos is 117.61s and the average length of the video segments is 36.18s.

\noindent\textbf{TACoS}~\citep{regneri2013grounding} consists of 127 videos of cooking activities. For a fair comparison, we follow the experimental setting in~\citep{gao2017tall}, using 10,146, 4,589, 4,083 moment-sentence pairs for training, validation and testing respectively. The average duration of the videos is 287.14s and the average length of the video segments is 5.45s. Compared with ActivityNet Captions, TACoS has a longer average duration of videos and a shorter average length of temporal moments described by language queries, which is more challenging.

\noindent\textbf{Evaluation Metrics.} Following~\citep{gao2017tall}, we adopt ``R@$n$, IoU=$m$'' to evaluate the model's performance. It is deﬁned as the percentage of language queries having at least one correct moment whose IoU with ground truth is larger than $\theta$ in the top-n retrieved moments. As PRVG regresses one moment for each query, we set $n$ = 1, $\theta \in$ \{0.3, 0.5, 0.7\} for ActivityNet Captions and $\theta \in \{0.1, 0.3, 0.5\}$ for TACoS. And we also use the “mIoU” metric which calculates the average IoU with ground truth over all testing samples to compare the overall performance.

\subsection{Implementation Details}
We use Glove word2vec~\citep{pennington2014glove} to generate word embedding vector, and feed the sequence of word embedding into a three-layer LSTM network~\citep{hochreiter1997long} to get the feature representations of sentences. For a fair comparison, we use the same visual features extracted by pre-trained C3D~\citep{tran2015learning} on ActivityNet Captions and TACoS. The channel numbers of sentence feature and video feature are all set to 512. The number of sampled clips N is set to 256, 512 for ActivityNet Captions and TACoS, respectively. \par

We use Adam~\citep{kingma2014adam} for optimization. The learning rate is set to 0.0001 for ActivityNet Captions and TACoS. The batch size is set to 32 for ActivityNet Captions and 16 for TACoS respectively. The trade-off parameters $\lambda$ and $\beta$ are set to 2 and 2 for both ActivityNet Caption and TACoS.
We adopt the same split for both datasets and sampling strategy as~\citep{bao2021dense} for training and inference. Specifically, we randomly sample $K (2 \leq K \leq 8)$ sentence queries with temporal order for each training sample. During inference, we first split annotated paragraph queries with more than 8 sentences into multiple sub-paragraphs with lengths not exceeding 8, and then take the whole paragraph as input.

\begin{table*}[t]
\centering
\scalebox{1.0}{
\begin{tabular}{c|cc|cc|cc}
\hline
\multirow{2}{*}{Row\#} & \multicolumn{2}{c|}{Train} & \multicolumn{2}{c|}{Test} & \multicolumn{2}{c}{mIoU}        \\
                       & Multiple     & Ordered     & Multiple    & Ordered     & ActivityNet Captions       & TACoS        \\ \hline
1                      & $\surd$      & $\surd$     & $\surd$     & $\surd$     & \textbf{55.62} & \textbf{29.18} \\
2                      & $\surd$      & $\times$    & $\surd$     & $\times$    & 47.21          & 22.99          \\
3                      & $\times$     & -           & $\times$    & -           & 44.69          & 22.44          \\
4                      & $\surd$      & $\surd$     & $\times$    & -           & 42.91          & 20.89          \\
5                      & $\surd$      & $\surd$     & $\surd$     & $\times$    & 29.06          & 14.61          \\ \hline
\end{tabular}
}
\caption{PRVG performance for different training and testing settings on ActivityNet Captions and TACoS.}
\label{table2}
\end{table*}

\subsection{Comparison with the State-of-the-Art Methods}
We evaluate the proposed PRVG model on ActivityNet Captions and TACoS, and compare it with recently proposed state-of-the-art methods of video grounding and dense video grounding. Video grounding methods can be grouped into two categories: 1) Proposal-based Approaches: CTRL~\citep{gao2017tall}, MCN~\citep{anne2017localizing}, TGN~\citep{chen2018temporally}, QSPN~\citep{xu2019multilevel}, 2D-TAN~\citep{zhang2020learning}, CBLN~\citep{liu2021context}, TACI~\citep{shin2022learning} and BMRN~\citep{Seol_2023_CVPR}. 2) Proposal-free Approaches: ABLR~\citep{yuan2019find}, DRN~\citep{zeng2020dense}, VSLNet~\citep{zhang2020span}, BPNet~\citep{xiao2021boundary} and CPNet~\citep{li2021proposal}. Dense video grounding methods include Beam Search~\citep{bao2021dense}, 3D Temporal-Paragraph Network (3D-TPN) (a extension of 2D-TAN implemented by~\citep{bao2021dense}), DepNet~\citep{bao2021dense}, SVPTR~\citep{Jiang_2022_CVPR} and HSCNet~\citep{Tan_2023_CVPR}. The results are reported in Table \ref{table1}. From the results, we can find that PRVG achieves comparable results with previous state-of-art methods, which demonstrates the effectiveness of our proposed PRVG model. \par

First, PRVG achieves superior results to proposal-based methods. We analyze that proposal-based approaches have to manually generate proposals first, which are usually with fixed length and heavily rely on prior knowledge, leading to the incapability to produce more accurate results. In contrast, our PRVG directly predicts temporal boundaries of moments of arbitrary duration and thus can get more accurate localization results. \par

In addition, we compare PRVG with proposal-free methods. Experimental results show that PRVG works better than most proposal-free methods. We analyze that their inferior performance might come from the dense prediction in nature. They typically predict much more than one temporal moment for one language query, and thus need to carefully design sophisticated strategies to distinguish them. Different from these approaches, our PRVG directly predicts temporal boundaries in a one-to-one manner by using languages as queries. \par

Finally, we compare PRVG with other dense video grounding methods. PRVG outperforms the Beam Search method. Except for Rank1@0.5 on TACoS, PRVG achieves much better performance than 3D-TPN and DepNet on all evaluation metrics for TACoS and ActivityNet Captions. This is because 3D-TPN is incapable of modeling the interactions among sentences in a paragraph effectively, and DepNet loses fine-grained context information for a specific language query when aggregating temporal information of multiple sentence queries into a compact set. However, our PRVG effectively extracts context information without information compression by modeling pair-wise interactions with self-attention. \textcolor{black}{And PRVG performs similarly to SVPTR, while worse than HSCNet. We analyze that the reasons may be that HSCNet introduces larger computation complexity (3 layers and 9 layers for ActivityNet Captions and TACoS respectively, while 1 layer in our PRVG), stronger multi-grained crossmodal interaction (hierarchical designs), and more training loss for multimodal alignment (encoder and decoder loss).}

\begin{table}[t]
\centering
\scalebox{1.0}{
\begin{tabular}{c|cccc}
\hline
Method           & 0.1   & 0.3   & 0.5   & mIoU  \\ \hline
PRVG (bbox)      & 60.82 & 40.14 & 22.54 & 26.47 \\
PRVG (bbox + pw) & 59.96 & 40.39 & 23.54 & 27.17 \\
PRVG (bbox + pl) & \textbf{61.64} & \textbf{45.40} & \textbf{26.37} & \textbf{29.18} \\ \hline
\end{tabular}

}
\caption{Ablation study on proposal-level attention loss on TACoS. 'bbox', 'pw' and 'pl' mean bounding box loss, position-wise attention loss and proposal-level attention loss respectively.}
\label{table3}
\end{table}

\begin{table}[t]
\centering
\begin{tabular}{c|cccc}
\hline
Method          & 0.3            & 0.5            & 0.7            & mIoU           \\ \hline
Average Pooling & 78.05          & 60.71          & 37.23          & 55.34          \\
TSN            & \textbf{78.27} & \textbf{61.15} & \textbf{37.83} & \textbf{55.62} \\ \hline
\end{tabular}
\caption{Ablation study on sampling methods for video clips on ActivityNet Captions.}
\label{table4}
\end{table}

\subsection{Ablation Studies}
\noindent \textbf{Study on the dense video grounding setting.}
To verify the effectiveness of dense video grounding setting, i.e., inputting multiple temporally ordered sentences in a paragraph as queries, we train and test PRVG in different settings: 1) one sentence or multiple sentences as input. 2) the input sentences are temporally ordered or not. The results in Table~\ref{table2} (Row 1-3) demonstrate that dense video grounding setting is helpful for more accurate video grounding, since the semantic relevance and temporal cues of multiple temporally ordered sentences in a paragraph can provide more temporal context to avoid semantics ambiguity.

\begin{table}[t]
\centering
\small\scalebox{1.0}{
\begin{tabular}{c|cc|c}
\hline
\multirow{2}{*}{Method} & \multicolumn{2}{c|}{mIoU} & Test time (s) \\
                        & VG         & Dense VG     & Dense VG      \\ \hline
DETR-VG                 & 37.67      & 37.68        & 507           \\
Language DETR-VG        & 41.50      & 41.27        & 513           \\
PRVG                    & \textbf{44.69}      & \textbf{55.62}        & \textbf{227}           \\ \hline
\end{tabular}
}
\caption{Performance and test time comparison with DETR-based models on ActivityNet Captions. PRVG outperforms DETR-VG and Language DETR-VG on both mIoU and inference speed.}
\label{table5}
\end{table}

\begin{figure}[t]
\centering
\includegraphics[width=1.0\columnwidth]{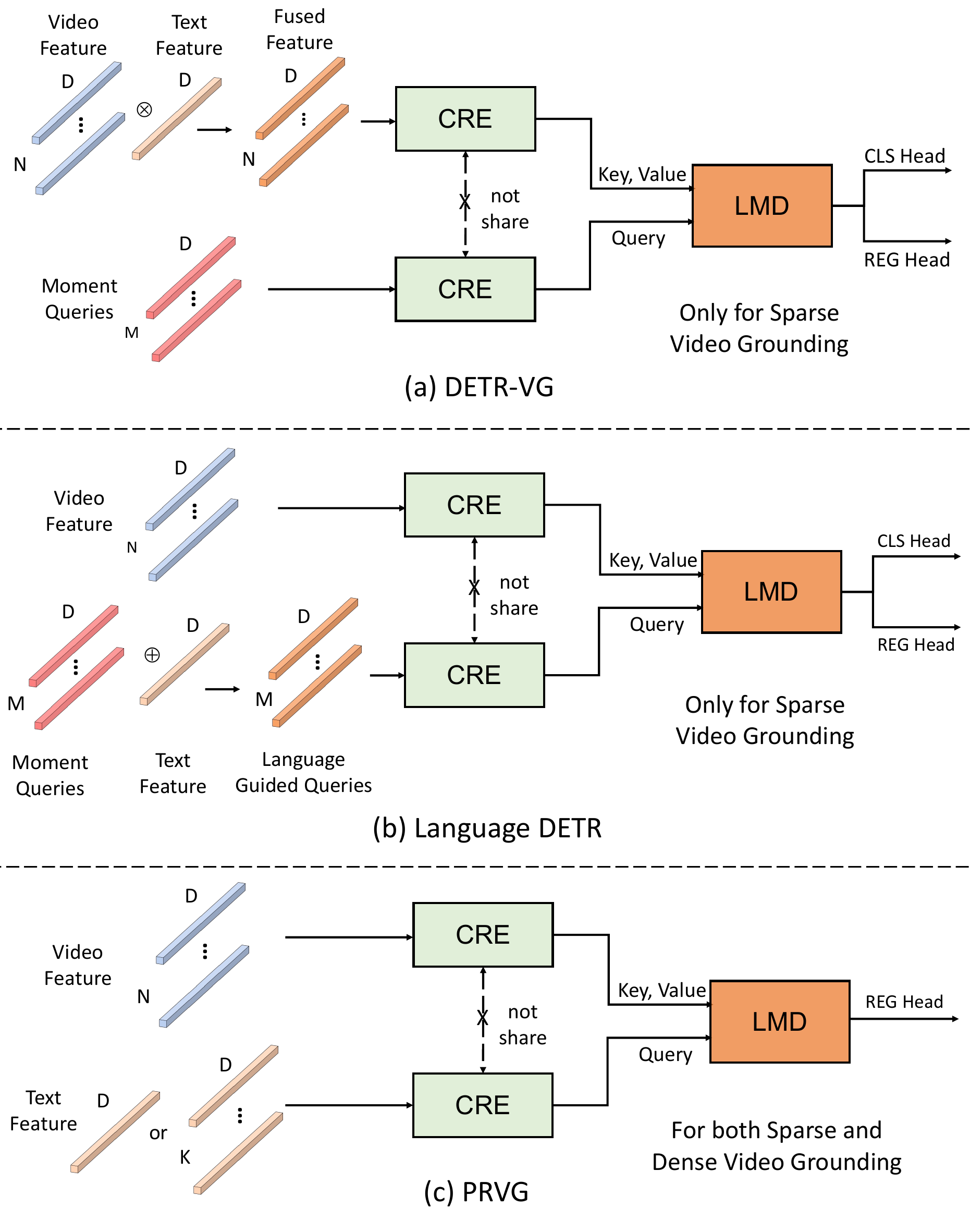}
\caption{The illustrations of two DETR-based video grounding methods and our proposed PRVG. (a) DETR-VG. (b) Language DETR. (c) PRVG. DETR-VG first performs element-wise multiplication for multi-modal fusion, and then use fixed number of learnable moment queries to decode the temporal boundary of the language description. Language DETR injects text features into the moment queries, and uses the fused language guided queies to decode the temporal boundary. While our PRVG uses language as queries for parallel regression. By removing the classification branch, PRVG can directly predict the temporal boundary for each language description, which gets rid of the complicated label assignment and post-processing. Moreover, PRVG accepts any number of language descriptions at once, thus can deal with both sparse and dense video grounding. Limited by the fixed number of moment queries, the other two methods can only perform sparse video grounding, and need time-consuming beam search for dense video grounding.}
\label{fig5}
\end{figure}

Meanwhile, the result in Table \ref{table2} (Row 3) shows that PRVG designed for dense video grounding task can achieve comparable performance to the state-of-the-art video grounding methods on ActivityNet Captions dataset in sparse video grounding setting,i.e. inputting one sentence query for both training and testing. From the results in Table \ref{table2} (Row 4), even if trained with multiple temporally ordered sentences, our PRVG still gets good results for testing with one sentence query, which further demonstrates the generalization ability of PRVG. We also test PRVG trained by multiple temporally ordered language queries with multiple temporally independent (disordered) language queries as the input of decoder. As shown in Table \ref{table2} (Row 5), the performance degradation is evident, demonstrating that PRVG can capture the temporal interactions among these temporally ordered sentences during training phase.

\noindent \textbf{Study on the proposal-level attention loss.}
Compared with ActivityNet Captions, TACoS has a longer average duration of videos and a shorter average length of temporal moments described by language queries, which is more challenging. Therefore, we conduct ablation study on our proposed proposal-level attention loss on TACoS. We re-train our model with the following settings: 1) PRVG (bbox): trained with only temporal boundary regression loss. 2) PRVG (bbox + pw): trained with temporal boundary regression loss and position-wise attention loss~\citep{yuan2019find}. 3) PRVG (bbox + pl): trained with temporal boundary regression loss and our proposed proposal-level attention loss. Table \ref{table3} shows the performance comparisons. Adding these two attention losses to bbox loss both improve the performance of PRVG, but our proposed proposal-level attention loss achieves better results, which demonstrates the effectiveness of the proposal-level attention loss. More in-depth analysis of proposal-level attention loss is described in section \ref{Qualitative_Analysis}.

\noindent \textbf{Study on sampling methods for video clips.} We try two sampling methods to reduce the number of video clips for fast processing, including average pooling and TSN sampling~\citep{wang2016temporal}. For average pooling, we aggregate the features of adjacent video clips into one feature. For TSN, we uniformly sample one clip for one segment. We sample 512 clips for both methods. Table \ref{table4} shows that PRVG with TSN sampling achieves better performance, which we conjecture that average pooling may over-smooth the temporal information and random sampling in TSN can be regarded as a kind of data augmentation. Therefore, we adopt TSN as the default sampling method.

\begin{figure*}[t]
\centering
\includegraphics[width=0.85\textwidth]{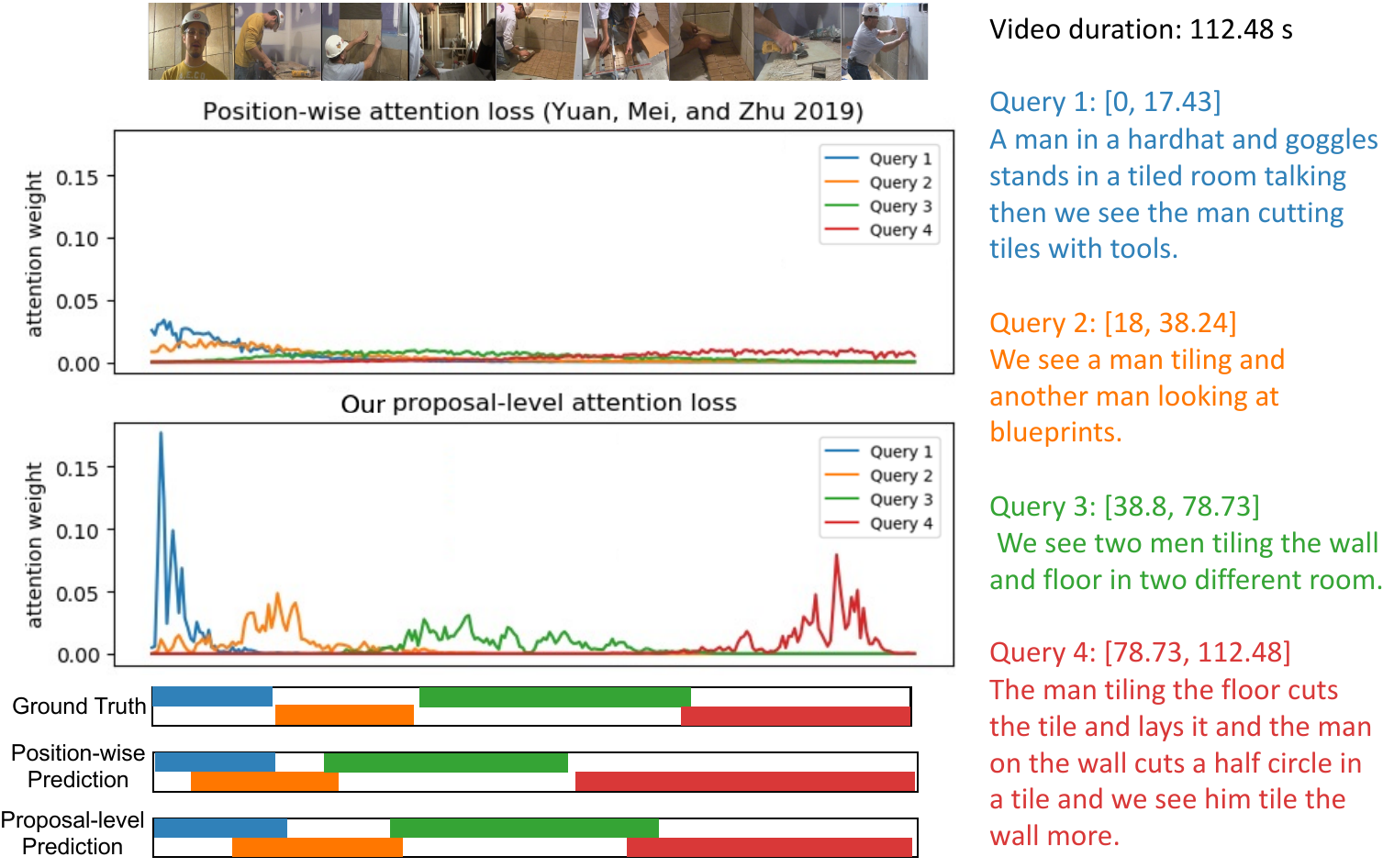} 
\caption{Visualization of decoder's attention distribution and dense video grounding results on the ActivityNet Captions. After training with our proposed proposal-level attention loss, PRVG distinguishes between foreground and background better, and focuses on the corresponding regions for all queries more accurately. Moreover, the multimodal attention distribution indicates our proposal-level attention loss can further enhance positions that are useful for boundary regression in the foreground region.
}
\label{fig6}
\end{figure*}

\noindent \textbf{Comparison with DETR-based models.}
\label{section:detr}
To compare with DETR-based models and verify the effectiveness of our design of using languages as queries, we devise two baseline models with DETR~\citep{carion2020end}, marked as DETR-VG and Language DETR-VG respectively. As shown in Fig.\ref{fig5}, both baseline models adopt the same architecture as the original DETR, and have 10 learnable moment queries similar to the object queries in the original DETR. Specifically, for DETR-VG, we first perform cross-modal fusion between video and sentence features by element-wise multiplication, and then input the fused feature into the Transformer. For Language DETR-VG, the input of the encoder is only the video feature, and it uses language queries as the initial input of the decoder. To tackle the dense VG task, we use beam search as post-processing method like~\citep{bao2021dense}. In particular, we first localize each sentence in the paragraph independently, then apply beam search on the top 3 grounding results of each sentence to make the final dense VG results match the temporal order.

Table \ref{table5} shows that the performance of Language DETR-VG is better than DETR-VG by using language queries as initial input of the decoder. Using languages as queries further, PRVG outperforms both baselines by a large margin. 
On the one hand, the design of languages as queries makes each query in our decoder represent the concrete semantics of a sentence and we can supervise decoder's attention distribution to guide the model to attend the ground truth moment. 
On the other hand, since each language query in the decoder directly predicts the moment corresponding to its sentence description without classification and there are no negative examples, thus eliminating the imbalance of positive and negative examples and performing better. 

Besides, when extending from VG to dense VG, the performance of DETR-based models for dense VG is almost the same as that for VG, which shows that DETR-based models with beam search as post-processing cannot capture the temporal and semantic relationship among the sentences in the paragraph.
However, PRVG for dense VG fully exploits their temporal interaction and achieves much better performance than PRVG for VG. In addition, we compare the test time of three models on ActivityNet Captions for dense VG task. As shown in Table \ref{table5}, DETR-based models take more than twice the time of PRVG. This is due to the complicated and time-consuming post-processing procedure in DETR-based models. Via parallel regression, PRVG directly regresses the moment for each sentence without any post-processing technique, leading to more efficient inference. 

\subsection{Qualitative Analysis}
We visualize an example to illustrate whether PRVG can attend to the ground truth and compare our proposal-level attention loss (pl) with position-wise attention loss (pw). 
As shown in Fig. \ref{fig6}, PRVG trained with pl assigns more attention weights to the ground truth than background, and focuses on the corresponding regions for all queries accurately. 
However, for PRVG trained with pw, decoder's attention is much smoother and there is little difference between foreground and background.
Moreover, our attention in the ground truth presents a multi-peak shape, indicating that proposal-level attention loss can not only help PRVG attend to the region corresponding to the language description, but also determine the importance of different positions within this region, thus further enhancing positions that are useful for boundary regression.

\label{Qualitative_Analysis}

\section{Conclusion}
In this paper, we have proposed a simple yet effective end-to-end framework (PRVG) for dense video grounding, which directly regresses the temporal boundaries of moments by using a paragraph as input. The key in our design is to use languages as queries and directly regress one moment for each sentence without any post-processing technique. We also propose a robust and scale-invariant proposal-level attention loss to guide the training of PRVG. Extensive experiments on TACoS and ActivityNet Captions demonstrate the superior performance of PRVG and the effectiveness of parallel decoding paradigm.
\bibliographystyle{model2-names}
\bibliography{refs}

\end{document}